\documentclass[letterpaper, 10 pt, conference]{ieeeconf}  

\IEEEoverridecommandlockouts                              

\usepackage{amsmath,amsfonts}
\usepackage{algpseudocode}
\usepackage[linesnumbered,ruled,vlined]{algorithm2e}
\usepackage{array}
\usepackage[caption=false,font=scriptsize,labelfont=sf,textfont=md]{subfig}
\usepackage{textcomp}
\usepackage{stfloats}
\usepackage{url}
\usepackage{verbatim}
\usepackage{graphicx}
\usepackage{cite}
\usepackage{amssymb}
\usepackage[mathscr]{eucal} 
\usepackage{diagbox}
\usepackage{makecell}
\usepackage{xcolor}
\usepackage{units}
\usepackage{multirow}

\usepackage{booktabs}

\usepackage[capitalise]{cleveref} 
\let\labelindent\relax
\usepackage[inline]{enumitem}

\title{\LARGE \bf
PFEA: An LLM-based High-Level Natural Language Planning and Feedback Embodied Agent for Human-Centered AI
}

\author{
Wenbin Ding, Jun Chen$^{*}$, Mingjia Chen, Fei Xie, Qi Mao, and Philip Dames
\thanks{$^{*}$Corresponding author: jun.chen@nnu.edu.cn}
\thanks{W.~Ding, J.~Chen, M.~Chen, F.~Xie, and Q.~Mao are with the School of Electrical and Automation Engineering, Nanjing Normal University, Nanjing, Jiangsu 210023, China}
\thanks{P.~Dames is with the Department of Mechanical Engineering, Temple University, Philadelphia, PA 19122, USA}%
}

\begin{document}

\maketitle
\thispagestyle{empty}
\pagestyle{empty}

\begin{abstract}
The rapid advancement of Large Language Models (LLMs) has marked a significant breakthrough in Artificial Intelligence (AI), ushering in a new era of Human-centered Artificial Intelligence (HAI). HAI aims to better serve human welfare and needs, thereby placing higher demands on the intelligence level of robots, particularly in aspects such as natural language interaction, complex task planning, and execution. Intelligent agents powered by LLMs have opened up new pathways for realizing HAI. However, existing LLM-based embodied agents often lack the ability to plan and execute complex natural language control tasks online. This paper explores the implementation of intelligent robotic manipulating agents based on Vision-Language Models (VLMs) in the physical world. We propose a novel embodied agent framework for robots, which comprises a human-robot voice interaction module, a vision-language agent module and an action execution module. The vision-language agent itself includes a vision-based task planner, a natural language instruction converter, and a task performance feedback evaluator. Experimental results demonstrate that our agent achieves a 28\% higher average task success rate in both simulated and real environments compared to approaches relying solely on LLM+CLIP, significantly improving the execution success rate of high-level natural language instruction tasks.
\end{abstract}

\section{Introduction}

The advancement of AI has ushered in a new era of HAI. The rapid development of LLMs \cite{vaswani2017attention, brown2020language}, in particular, has accelerated the progress of industry 5.0, which prioritizes Human-centric Smart Manufacturing (HSM) as a foundational pillar \cite{wang2022toward}. This new paradigm is centered on human needs and well-being. The realization of this objective necessitates effective human-robot collaboration \cite{rabby2022learning}, requiring robotic systems to be equipped with advanced multimodal natural interaction capabilities, perception and environment understanding, cognitive reasoning, intelligent decision-making and task planning capabilities.

HAI is transforming traditional modes of working, living, and learning. Unlike conventional technology-driven AI, HAI places human subjectivity at the core of the intelligent ecosystem. By emphasizing human needs, values, and societal well-being, HAI seeks to empower people and deliver smarter, more convenient solutions through AI. This shift marks a transition from a technology-centered paradigm to one that is centered on humans and society. In recent years, the perception, reasoning, and interaction capabilities of intelligent robots play a crucial role in the success of HAI systems. However, with the high demands of HAI, further exploration and breakthroughs are still needed in achieving more natural instruction execution and embodied closed-loop control (reviewed in \cref{sec:II_A}).

LLMs have been trained with rich semantic knowledge of the real world, so can robots perform specific tasks in the real world? This leads to the key problem of how robots can perform human-oriented service tasks in specific environments based on high-level language instructions. 
In recent years, approaches to similar tasks, such as SayCan \cite{ahn2022can} and Grounded Decoding (GD) \cite{huang2023grounded}, have adopted a planner-executor architecture. However, these methods often lack adaptability to new environments and fail to incorporate feedback during task execution (reviewed in \cref{sec:II_B}).
To solve this problem, we propose an LLM-based high-level natural language \textbf{P}lanning and \textbf{F}eedback \textbf{E}mbodied \textbf{A}gent (PFEA).
We argue that effective task execution in embodied agents requires three key capabilities. First, the robot must be able to perceive and comprehend natural language instructions and identify task steps based on objects present in the environment. Second, it must organize and sequence these steps appropriately in accordance with the current environmental context. Finally, the robot should be equipped to receive and utilize task execution feedback to improve performance and success rates.

In this paper, we make the following contributions.
\begin{enumerate*}[label=\arabic*.]
\item We propose a vision-language-based unified scene understanding framework for robotic task perception and planning. This system integrates visual information from the environment into the task planner, enabling the decomposition of high-level language instructions into executable low-level actions tailored to the real-time environment.
\item We introduce a feedback control mechanism for task execution. By leveraging task-specific descriptions within concrete scenarios, the system evaluates whether a task has been successfully completed and feeds this information back to the planner, thereby enhancing the overall task success rate.
\item We construct an embodied agent system deployed in both simulated and real-world scenarios. Our embodied agent system establishes a direct connection between large models trained on virtual data and physical robots. We enable model deployment for robot control through a simple and effective, training-free approach, allowing seamless integration of large-scale models into real-world robotic systems.
\item We design a series of experiments that involve variations in object types, task categories, and the presence or absence of prompts to evaluate our system across multiple dimensions. We use these experiments to perform an ablation study of our system and to compare our system with baseline algorithms.
\end{enumerate*}

\section{Related Work}

\subsection{LLMs for Human-Centered AI}
\label{sec:II_A}
HAI primarily aims to place human subjectivity at the core of intelligent ecosystems.
Recent advancements are steadily enhancing the development of HAI, marking significant steps toward its seamless integration. Xu et al. \cite{xu2025embodied} emphasized the development and application of human-centered embodied agents. However, their implementation still faces challenges such as difficulties in long-term task planning.
Dorbala et al. \cite{dorbala2023can} proposed language-guided exploration, leveraging large language models to navigate previously unseen environments toward uniquely described target objects. Similarly, Wang et al. \cite{wang2024llm} introduced a vision-and-language collaborative robotic navigation approach to assist workers in retrieving tools. These studies illustrate how large models can empower embodied agents and accelerate the deployment of human-centered intelligent systems in real-world contexts. 
Leveraging the reasoning and generalization capabilities of large language models, our research further focuses on the deep integration of human needs, semantic intent, and environmental perception. This integration enables embodied agents to better understand and respond to high-level natural language instructions in real-world scenarios, thereby constructing embodied systems with enhanced adaptability and interactive intelligence.

\subsection{Vision and Language Embodied Agents}
\label{sec:II_B}

With the rise of agent-based LLMs, the field of embodied intelligence has seen rapid development. The core objective is to enable deep collaboration between LLMs and external tools, allowing LLMs not only to understand and generate language but also to manipulate physical-world robots to perform tasks~\cite{shridhar2022cliport, reed2022generalist, brohan2023rt}. In this way, these models are essentially given “hands and feet,” shifting from being able to “speak” to being able to “act.”

Embodied control based on large models can be broadly categorized into two types according to the way the LLMs is trained: end-to-end large-model embodied control~\cite{brohan2022rt} and agent-based hierarchical control using large models~\cite{vemprala2024chatgpt}. The former involves training LLMs on datasets comprising language, vision, and robotic actions~\cite{zitkovich2023rt, wen2025tinyvla, kim2024openvla}. However, this approach often struggles with limited data and poor generalization across diverse real-world environments~\cite{cai2024spatialbot}. The latter leverages hierarchical architectures within embodied agents to address robotic perception, decision-making, and control without the need for costly retraining of the LLMs. This makes it a low-cost yet effective and reliable alternative~\cite{liu2024ok, hu2023look}. The emergence of agent systems like MetaGPT~\cite{hong2023metagpt} has laid the groundwork for extending such agents into the physical world.

Compared to training specialized large models for direct embodied control, agent-based approaches offer lower cost and generally better performance. Therefore, this paper focuses on embodied control through agent-based systems that utilize large language models.

To address the challenge of enabling embodied agents to intelligently execute tasks, Ahn et al.\cite{ahn2022can} first proposed integrating LLMs with pretrained robotic skills and aligning them with real-world tasks via value functions, allowing robots to understand and carry out natural language instructions. Building upon this, Huang et al.\cite{huang2023grounded} introduced the GD method. While these systems, such as SayCan and GD, have made progress, they commonly lack feedback mechanisms during execution. In our work, we introduce a high-level natural language planner augmented with visual environmental perception, and further incorporate a task execution feedback module. This enables the agent to perceive execution outcomes and adjust its strategies and actions accordingly, thereby improving task success rates in complex and dynamic environments.

\section{PFEA}

\begin{figure*}[tbp]
\centering
\includegraphics[width=2\columnwidth]{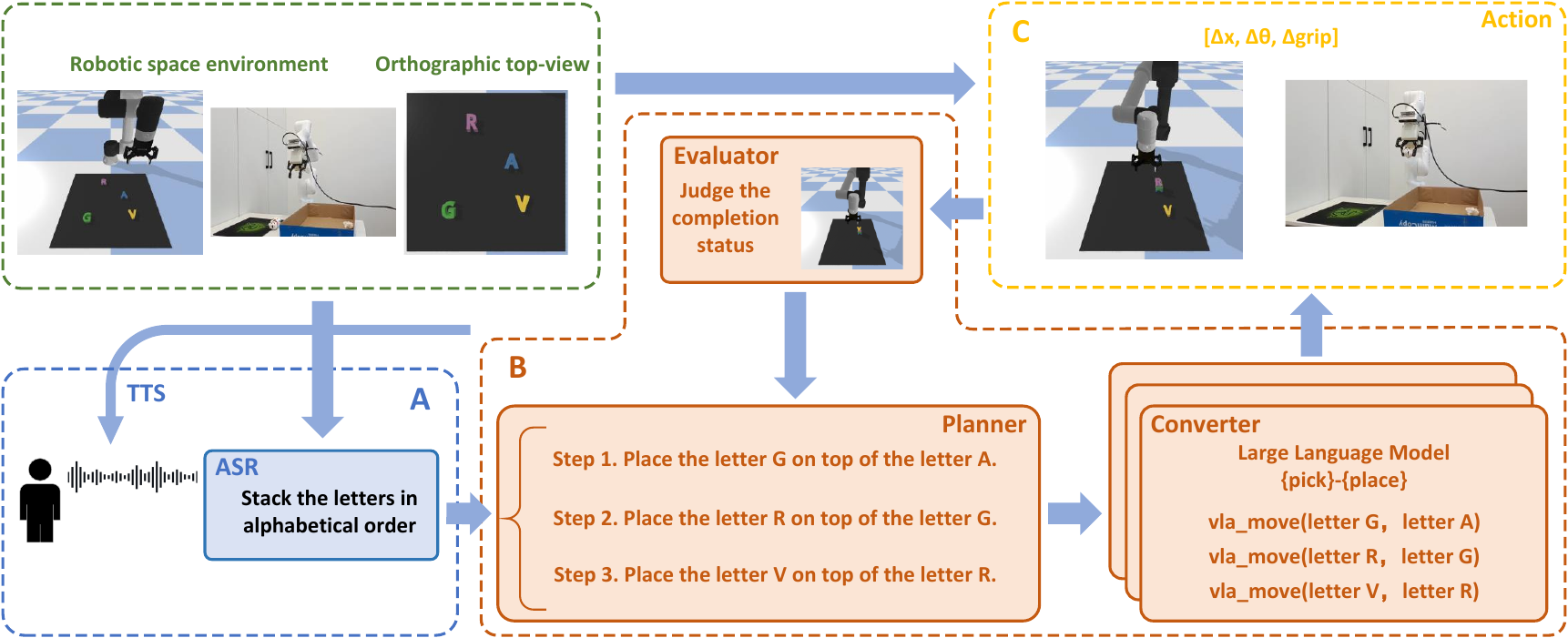}
\caption{
Overall system architecture of PFEA. The complete workflow begins with human voice input, followed by the agent performing speech recognition, vision-language task planning, conversion into executable python instructions, task execution through robotic actions, and task completion assessment. Finally, the agent generates a natural language response reporting the execution result.
}
\label{fig:VLEA_frame}
\end{figure*}

The overall agent in this paper is illustrated in Figure~\ref{fig:VLEA_frame}.
The LLM-based vision and language embodied agent system consists of three components, marked with A, B and C in Figure~\ref{fig:VLEA_frame}. The first component focuses on the agent's information conversion for direct interaction with humans. This component converts speech into text for the agent to understand, and then turns the agent’s responses into speech, enabling natural communication. 
The second component is a VLM-based vision and language agent, which primarily consists of three modules: a planner, a converter, and an evaluator. The planner decomposes high-level human commands into multi-step executable instructions. The converter extracts key information from these commands to generate robot control instructions. During task execution, the evaluator verifies and provides feedback on task completion in both simulated and real-world environments.
The third component is robot action execution, which detects and localizes real objects in image frames by their semantic meanings utilizing open vocabulary models, allowing the robot to plan a path to execute grasp actions. 

\subsection{Speech Information Processing}
To facilitate direct interaction with humans, our embodied agent is equipped with capabilities analogous to human ``hearing'' and ``speaking." This allows users to communicate with the robot using natural spoken language. The embodied agent receives human speech input, which is first processed by an Automatic Speech Recognition (ASR) \cite{wu2023decoder} model to convert the audio into plain text. This text is then passed to the agent’s planner for high-level language-based planning and execution, effectively completing the ``hearing" function of the robot. This enables the embodied system to further integrate into the real physical world.

Once the agent has received a command and formulated a plan, it executes the planned task step-by-step. After each step, the agent provides verbal feedback describing the action taken and the outcome, thus realizing the ``speaking" capability. This feedback is generated using a Text-To-Speech (TTS) \cite{jiang2023mega} model, enabling the robot to respond in a manner that is conversational like a human.
For ASR and TTS, we use the following models trained within the FunAudioLLM framework \cite{an2024funaudiollm}: SenseVoice to perform multidimensional voice analysis and CosyVoice to synthesize highly natural speech.

To initiate audio capture, the agent uses a noise-trigger mechanism. During the initial one-second monitoring window, it continuously samples input from the microphone. Recording is triggered once the input audio level exceeds a predefined noise threshold. The capture ends when the detected volume remains below the threshold for two consecutive seconds, ensuring precise endpoint detection of spoken utterances.

\subsection{Vision-Language Agent}

\begin{figure*}[tbp]
\centering
\subfloat[Planner Prompt]{
	\includegraphics[width=0.65\columnwidth]{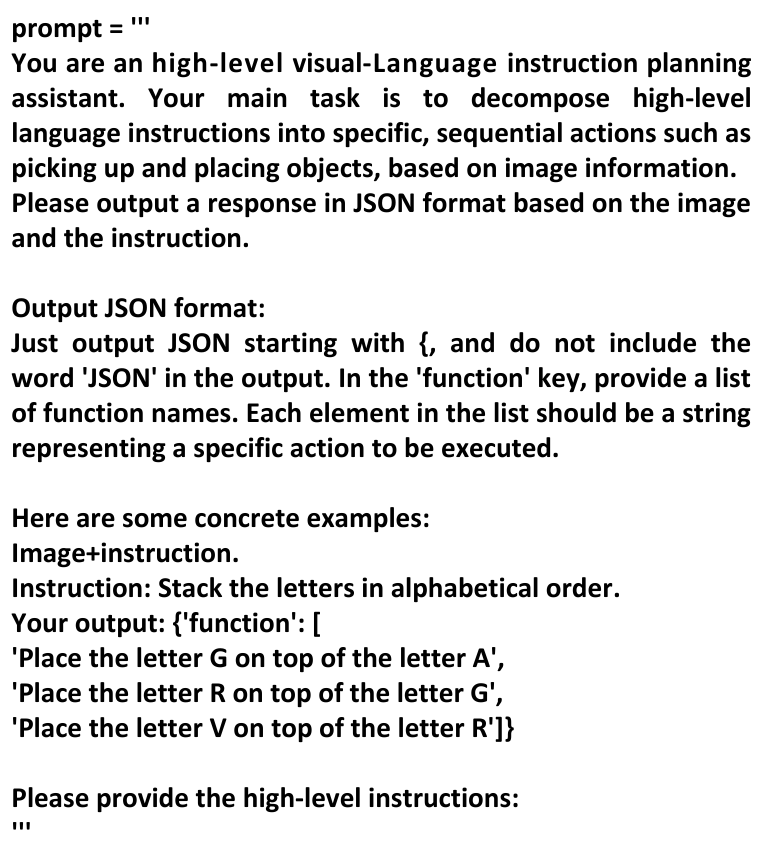}
	\label{fig:Planner}
}
\subfloat[Converter Prompt]{
	\includegraphics[width=0.65\columnwidth]{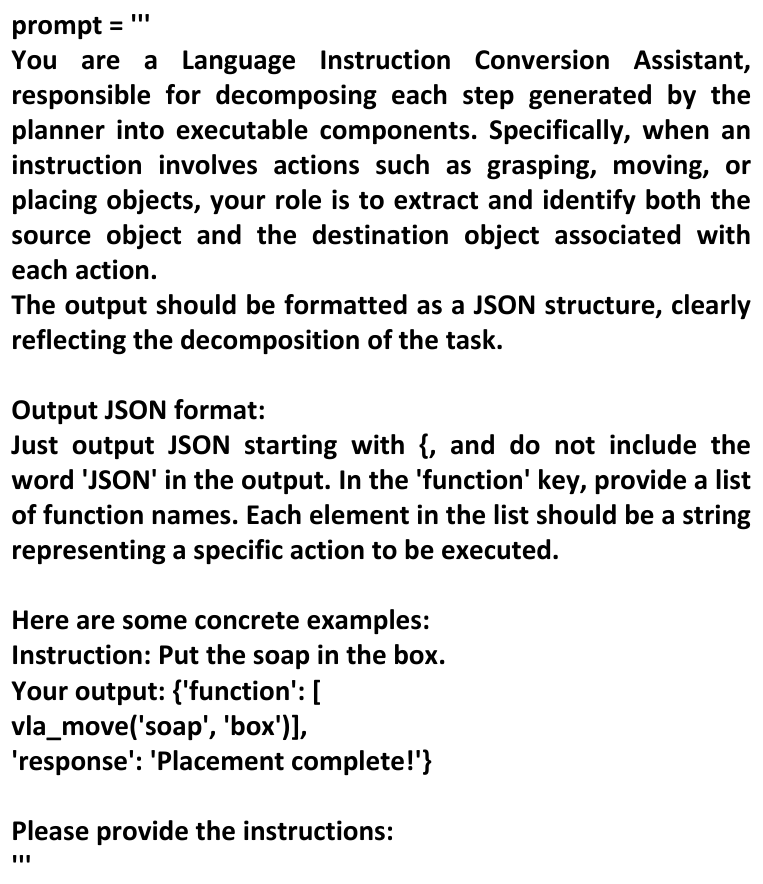}
	\label{fig:Converter}
} 
\subfloat[Evaluator Prompt]{
	\includegraphics[width=0.65\columnwidth]{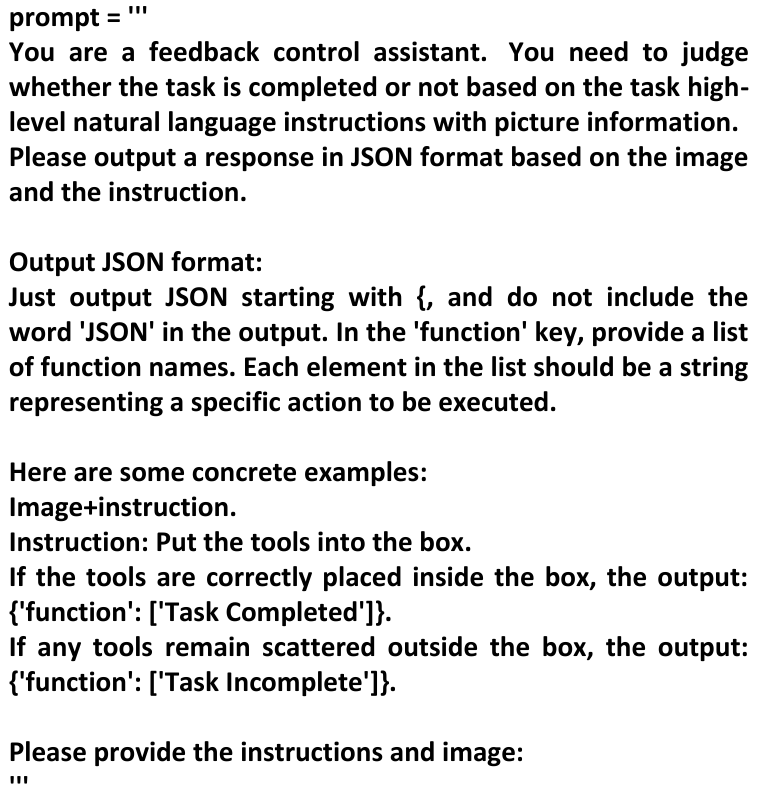}
	\label{fig:Evaluator}
}
\caption{
The prompts for the planner, converter, and evaluator are shown in Figure~\ref{fig:Planner}, Figure~\ref{fig:Converter}, and Figure~\ref{fig:Evaluator}, respectively.
}
\label{fig:prompt}
\end{figure*}

This component represents the core of our embodied intelligence system and functions analogously to the “brain” of the robot. 

\subsubsection{Planner}
Upon receiving high-level instructions from a human user, the agent first processes the command using a planner based on VLMs. This planner interprets the abstract human instruction in context and decomposes it into a sequence of specific and executable subtasks.
We adopt a VLM-based planner instead of relying solely on an LLMs and grounding model, as realized in systems like SayCan \cite{ahn2022can}. This is motivated by the necessity of tight coupling between the task execution and the real-world environment, which cannot be achieved solely by relying on the training distribution of grounding models. 
Our system primarily uses ChatGLM \cite{glm2024chatglm} as the LLM, though the system is also compatible with other models such as DeepSeek~\cite{liu2024deepseek, guo2025deepseek}, ChatGPT \cite{brown2020language, ouyang2022training}, and Gemini~\cite{team2023gemini}.

In practice, we design a structured prompt for the agent’s planner (see Figure~\ref{fig:Planner}), which includes a task description, relevant examples, and user input. The planner then decomposes the high-level instruction into multiple concrete execution steps, and outputs these steps to enhance system transparency and help users understand how the agent solves the task.

\subsubsection{Converter}
Given the planner, the robot still cannot execute the steps directly. Each subtask is passed to a converter module powered by ChatGLM \cite{glm2024chatglm}, which translates the linguistic instructions into executable robot control code. These final executable actions enable the robot to perform spatial manipulation tasks such as picking and placing objects.
We construct dedicated prompts for the agent's converter (see Figure~\ref{fig:Converter}), which guide the LLMs to convert each step (starting from the first) into executable python code, corresponding to the \textit{vlamove} skill library in the robotic grasping and action execution section. These commands are then carried out by the system’s action module.

\subsubsection{Evaluator}
Once all steps have been executed, the system invokes a VLM-based evaluator module to assess the outcome of the task.  After completing all steps, the evaluator determines task success and whether re-execution is necessary. If the execution is unsatisfactory, the system loops back to the planning phase to re-plan and reattempt the task.
It is important to emphasize that the design of the evaluator prompt should follow a result-oriented assessment strategy rather than directly checking whether the original task description has been fulfilled. This is because the post-execution visual scene may differ significantly from the initial state, making it unreliable to assess task success solely based on the original instruction.
The results from the evaluator are fed back to both the planner and the converter, enabling continual optimization of the system in terms of correctness and efficiency. Figure~\ref{fig:Evaluator} illustrates example task descriptions used in the evaluator prompts.

\subsection{Robotic Grasping and Action Execution}
This component serves as the physical execution core of the embodied agent, forming a closed loop between semantic instruction and real-world manipulation through integrated visual perception and motion control.
In this work, we adopt open-vocabulary models \cite{radford2021learning, minderer2023scaling} to construct a semantic-to-spatial mapping framework. This enables the agent to combine generated semantic parameters with RGB-D visual data to establish a dynamic understanding of its environment. The semantic features produced by the agent converter are cross-modally matched with real-time visual features to identify the spatial location of the target objects in the image. These 2D pixel coordinates are then transformed into 3D world coordinates using a calibrated projection matrix, achieving nonlinear mapping between image space and physical space.

During the grasping process, an RGB-D camera continuously captures real-time images. The model detects the target object and aligns coordinates accordingly. The robotic arm uses inverse kinematics algorithms to compute motion trajectories. Optimal joint angles are determined via Jacobian matrix solutions, allowing the manipulator to move precisely to the target position. Upon arrival, the gripper opens to grasp the object.
After a successful grasp, the robotic arm moves to the designated placement location, opens the gripper, and releases the object. Finally, the manipulator returns to its home position and waits for the next instruction.

\section{Experiments}
Our research employs a two-stage experimental design. Initially, a simulation experiment is conducted to determine whether PFEA can interpret high-level natural language commands, autonomously decompose them into a series of steps, and successfully perform tasks such as desktop organization. The core value of the simulation lies in its controllable environmental variables and high reproducibility, which allow for stable assessments of the system’s effectiveness and generalization ability under precisely defined conditions. This is difficult to achieve with physical hardware platforms, where complex and hard-to-regulate variables in real-world environments impede the precise validation of the system's core mechanisms.

Subsequently, building upon the simulation results, real-world experiments further assess the applicability of PFEA in practical scenarios. Physical hardware environments introduce real-world interactions and environmental disturbances that cannot be fully replicated in simulations. This allows for the validation of the system's adaptability under complex, real-world conditions. Consequently, this phase yields conclusions unattainable through simulation alone, specifically determining the practical efficacy of PFEA in performing tasks in genuine human environments. This serves as a critical supplement and the ultimate validation of the conclusions drawn from the simulation.

\subsection{Simulation Robotic Manipulation}

\begin{figure*}[tbp]
\centering
\includegraphics[width=2\columnwidth]{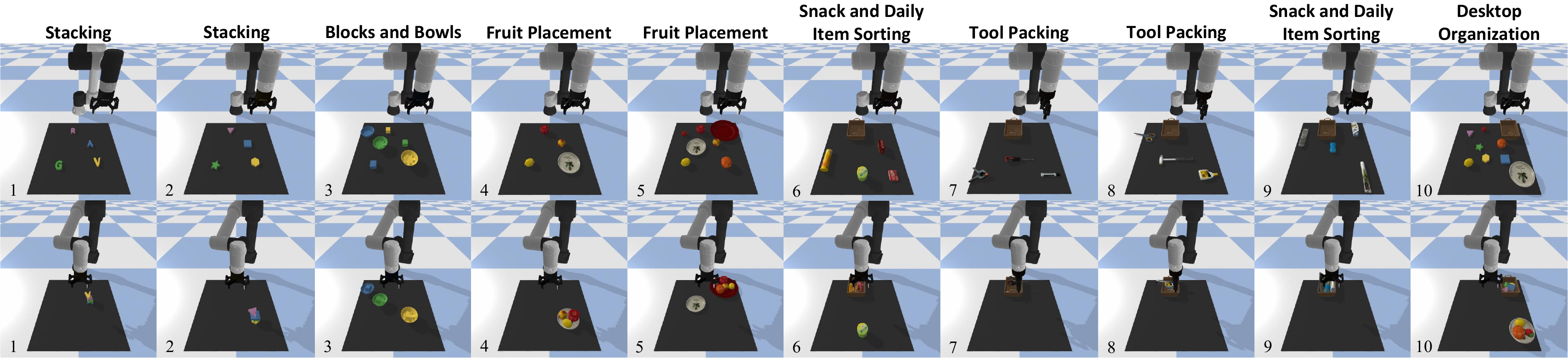}
\caption{
The first row shows the initial desktop scenarios corresponding to scenarios 1 to 10, and the second row shows the ten scenarios corresponding to the completion of the prompts listed in table \ref{table:sim_data}. The tasks in each column are categorized as shown in the figure.
}
\label{fig:10init_task}
\end{figure*}

\subsubsection{Simulation Setup}
We conduct experiments with a simulated tabletop manipulation environment based on RAVENS \cite{zeng2021transporter}.
The simulated setup includes an UR5e robotic arm, a Robotiq 2F-85 gripper, an Intel RealSense D435i RGB-D camera, and various tabletop objects, as illustrated in Figure~\ref{fig:10init_task}.
Based on these 10 environments, we developed a custom task suite consisting of 20 high-level natural language instruction tasks. Among them, 10 tasks include task-planning prompts, while the remaining 10 are unprompted, containing no prior examples in the agent's input. All 20 tasks are unseen, meaning the agent was not trained on them. Each task was tested 20 times.

The tasks are categorized as follows:
\begin{itemize}
  \item \textbf{Stacking}: Stacking letters or blocks according to alphabet or number of corners (``Stack the blocks in descending order of the number of corners").

  \item \textbf{Blocks and Bowls}: Rearranging blocks into color-matching bowls (``Put each block into the bowl of the same color").

  \item \textbf{Fruit Placement}: Placing all fruits into a designated dish (``Place all fruits into the red plate").

  \item \textbf{Snack and Daily Item Sorting}: Storing snacks and daily-use items into boxes (``Put all snacks except the chewing gum into the box").

  \item \textbf{Tool Packing}: Sorting and packing tools into appropriate containers (``Pack the tools into the box").

  \item \textbf{Desktop Organization (Mixed Category)}: Grouping different types of objects into separate containers for a clean desktop (``Put all fruits in the bowl and all blocks in the box").
\end{itemize}

Each of these tasks presents unique challenges. For instance, in stacking tasks, the formation of compound structures during early stages of stacking can lead to degraded visual representations. Three-dimensional overlaps between letter-shaped blocks often result in texture feature confusion, increasing the likelihood of task failure. To address this, we strategically place less interference-prone objects later in the stacking sequence.
Similarly, in fruit placement tasks, placing a large red apple first into a small bowl may block or push other fruits out due to spatial constraints. These examples highlight the critical role of step-wise planning in task success, underscoring 
the importance of the agent planner’s ability to prioritize about execution order.

\subsubsection{High-Level Task Execution}

\begin{table*}
\centering
\caption{Performance of PFEA in different experimental scenes and tasks. Metrics include planning success rate and overall task success rate. The first 10 tasks are prompted tasks, the last 10 tasks are unprompted tasks}
\label{table:sim_data}
\begin{tabular}{cllccc} 
\toprule
\multicolumn{1}{l}{}      &                                                  &                                                                               & PFEA                  & \multicolumn{1}{l}{PFEA} & LLM    \\ 
\cmidrule{4-6}
\multicolumn{1}{l}{Scene} & Type\textcolor[rgb]{0.106,0.133,0.173}{~of~}Task & Tasks                                                                         & Planning Success Rate & \multicolumn{2}{c}{Success Rate}  \\ 
\midrule
1                         & Stacking                                         & Stack the letter blocks in alphabetical order                                 & —                     & \textbf{90\%}            & 65\%   \\
2                         & Stacking                                         & Stack objects in descending order based on the number of corners          & —                     & \textbf{80\%}            & 60\%   \\
3                         & Blocks and Bowls                                 & Place each block into the bowl that matches its color                         & —                     & \textbf{90\%}            & 75\%   \\
4                         & Fruit Placement                                  & Place the fruits onto the plate                                               & —                     & \textbf{85\%}            & 70\%   \\
5                         & Fruit Placement                                  & Place all the fruits into the red plate                                       & —                     & \textbf{90\%}            & 65\%   \\
6                         & Snack and Daily Item                             & Put all the snacks into the box, excluding the chewing gum                    & —                     & 70\%                     & 70\%   \\
7                         & Tool Packing                                     & Put all the tools into the box                                                & —                     & \textbf{80\%}            & 60\%   \\
8                         & Tool Packing                                     & Organize the tools on the desk                                                & —                     & \textbf{80\%}            & 65\%   \\
9                         & Snack and Daily Item                             & Put all the daily-use items into the box                                      & —                     & \textbf{75\%}            & 70\%   \\
10                        & Mixed Category                                   & Place the fruits onto the plate and store the blocks in the box               & —                     & \textbf{80\%}            & 45\%   \\ 
\midrule
\multicolumn{1}{l}{}      &                                                  & \textbf{Prompted Tasks Total}                                                 & \multicolumn{1}{l}{}  & \textbf{82\%}            & 65\%   \\ 
\midrule
1                         & Blocks and Bowls                                 & Put the geometrically symmetric letters into the box                          & 45\%                  & \textbf{35\%}            & 5\%    \\
2                         & Stacking                                         & Stack the objects in ascending order based on the number of sides             & 65\%                  & \textbf{55\%}            & 10\%   \\
3                         & Blocks and Bowls                                 & Place three block into a bowl with a non-matching color                       & 90\%                  & \textbf{75\%}            & 65\%   \\
5                         & Fruit Placement                                  & Place some of fruits into the white plate and the rest into red plate & 90\%                  & \textbf{90\%}            & 15\%   \\
8                         & Tool Packing                                     & Organize the tools on the desk(replace a claw hammer)                         & 80\%                  & \textbf{70\%}            & 30\%   \\
9                         & Snack and Daily Item                             & Put all items on the desk into the box                                        & 85\%                  & \textbf{80\%}            & 10\%   \\
10                        & Blocks and Bowls                                 & Place the block with the most sides into the box                              & 95\%                  & \textbf{85\%}            & 75\%   \\
10                        & Mixed Category                                   & Place the fruits into the box and the blocks onto the plate                   & 95\%                  & \textbf{75\%}            & 70\%   \\
10                        & Fruit Placement                                  & Place only the fruits into the box                                            & 95\%                  & \textbf{90\%}            & 55\%   \\
10                        & Stacking                                         & Stack all the blocks sequentially onto the plate                              & 90\%                  & \textbf{85\%}            & 10\%   \\ 
\midrule
\multicolumn{1}{l}{}      &                                                  & \textbf{Unprompted Tasks Total}                                               & 83\%                  & \textbf{74\%}            & 35\%   \\
\bottomrule
\end{tabular}
\end{table*}

In this experiment, we considered four object categories commonly encountered in human-centered embodied AI scenarios: fruits, snacks, daily-use items, and tools. The agent is provided with high-level natural language instructions, such as "Place all the fruits into the red bowl" or "Organize the tools on the desk". Upon receiving the instruction, the agent first processes it through its planner module, while simultaneously acquiring RGB images from an onboard camera to support effective task planning.
The agent then sequentially executes each planned step. Through its task translator and action execution module, the agent identifies task-relevant start points and maps positions from image coordinates to real-world coordinates. The robotic arm is then actuated to carry out the planned manipulation actions.
After task execution, the agent invokes its task evaluator to determine whether the task has been successfully completed. If successful, the agent awaits the next instruction. If not, it re-engages the planner to revise the plan based on the current task state. Notably, the evaluator does not directly assess the initial high-level instruction. Instead, it uses a specially designed evaluation prompt tailored to the outcome of the task, focusing on goal-state verification.

\subsubsection{Unprompted High-Level Task Execution}
The following section presents experiments on unprompted tasks. 
These experiments were designed to achieve two primary objectives: first, to explicitly verify whether a VLM-driven agent can autonomously infer and execute the correct action sequence without relying on any step-by-step task prompts. Second, to rigorously evaluate the system’s generalization capability under conditions of minimal task-specific prior guidance.
In these experiments, our goal was to vary only the input high-level natural language commands while keeping all other components of the agent unchanged. The commands were deliberately designed to differ from the scenarios implicitly embedded in the base prompt template, ensuring that the agent could not rely on superficial prompt similarity to accomplish the tasks.

Therefore, we designed the following generalization experiments:
In stacking tasks, we entirely altered the expected stacking order, requiring the agent to automatically replan the sequence from scratch.
In the blocks and bowls task, color matching was intentionally removed, compelling the agent to formulate new object-to-container associations.
In fruit placement tasks, instructions were modified so that fruits needed to be distributed across two plates instead of one.
In the snack and daily item storing and tool packing tasks, previously unseen items were introduced—items never mentioned in any prompt—to test the agent's ability to plan and execute tasks involving entirely novel objects.
In the desktop organization task, the usual object-to-container mappings were reversed (e.g., fruits placed in boxes and blocks placed in bowls).
In a particularly challenging setup, we attempted to fuse multiple task types without any prior hint—for example, requiring the agent to perform block stacking inside a plate as part of the desktop organization task.
Experimental results show that the proposed agent excels in prompt-free generalization scenarios, achieving a success rate nearly twice that of the baseline approach using LLM+CLIP. This strong performance underscores the PFEA’s significant advantage in zero-shot generalization. Detailed success rates are presented in Table~\ref{table:sim_data}.

\subsubsection{Comparisons to Baselines}
To further validate the effectiveness of our proposed agent architecture, we conducted comparative experiments in a simulated environment. 
In these experiments, the baseline method was selected in strict accordance with the core functional modules of the SayCan \cite{ahn2022can} architecture, namely the collaborative integration framework of LLM and CLIP \cite{radford2021learning, minderer2023scaling}.
Figure~\ref{fig:10init_task} presents visual results of completed tasks. Experimental results show that the agent can reliably perform language understanding, task planning, task translation, action execution, and task evaluation, ultimately completing the instructed tasks and providing voice feedback upon success.
The per-task success rates are shown in Table~\ref{table:sim_data}. 
The performance of PFEA was evaluated across various experimental scenarios and tasks. The first 10 tasks are prompt-based, while the latter 10 are non-prompt tasks. Experimental scenarios 1 through 10 correspond to the ten environments shown in Figures \ref{fig:10init_task}, arranged from left to right and top to bottom. Evaluation metrics include task planning success rate and overall task completion rate. The task planning success rate refers to the success rate of PFEA's planner in generating plans for non-prompt tasks, serving as a measure of generalization ability. The task completion rate compares the success rates of PFEA and a baseline method using LLM and CLIP. Values with higher success rates are highlighted in bold.
Compared to baseline methods, the experimental results demonstrate that our proposed full framework is critical to successful task execution. PFEA achieved approximately a 28\% improvement in success rate compared to the LLM and CLIP method. In particular, the visual perception module plays a central role in enabling informed and reliable decision-making, while the evaluation and feedback mechanism is essential for completing complex task workflows.

Finally, the experimental results show that the proposed agent significantly outperforms the baseline that relies solely on large language model and CLIP, achieving a 28\% higher average task success rate. This improvement underscores the importance of each component within the agent—especially the visual perception, task planning, and evaluation-feedback mechanisms—which work in concert to support the execution of complex tasks.
The agent demonstrates strong performance not only in prompted environments but also in a range of challenging, practice-free, human-centered manipulation tasks. These include accurately classifying and placing objects with diverse categories and attributes into their appropriate containers, reflecting a solid understanding of task requirements and reliable action execution. This provides strong validation of the effectiveness of the agent we designed.

\subsection{Real-world Robotic Manipulation}

\begin{figure}[htbp]
\centering
\includegraphics[width=1\columnwidth]{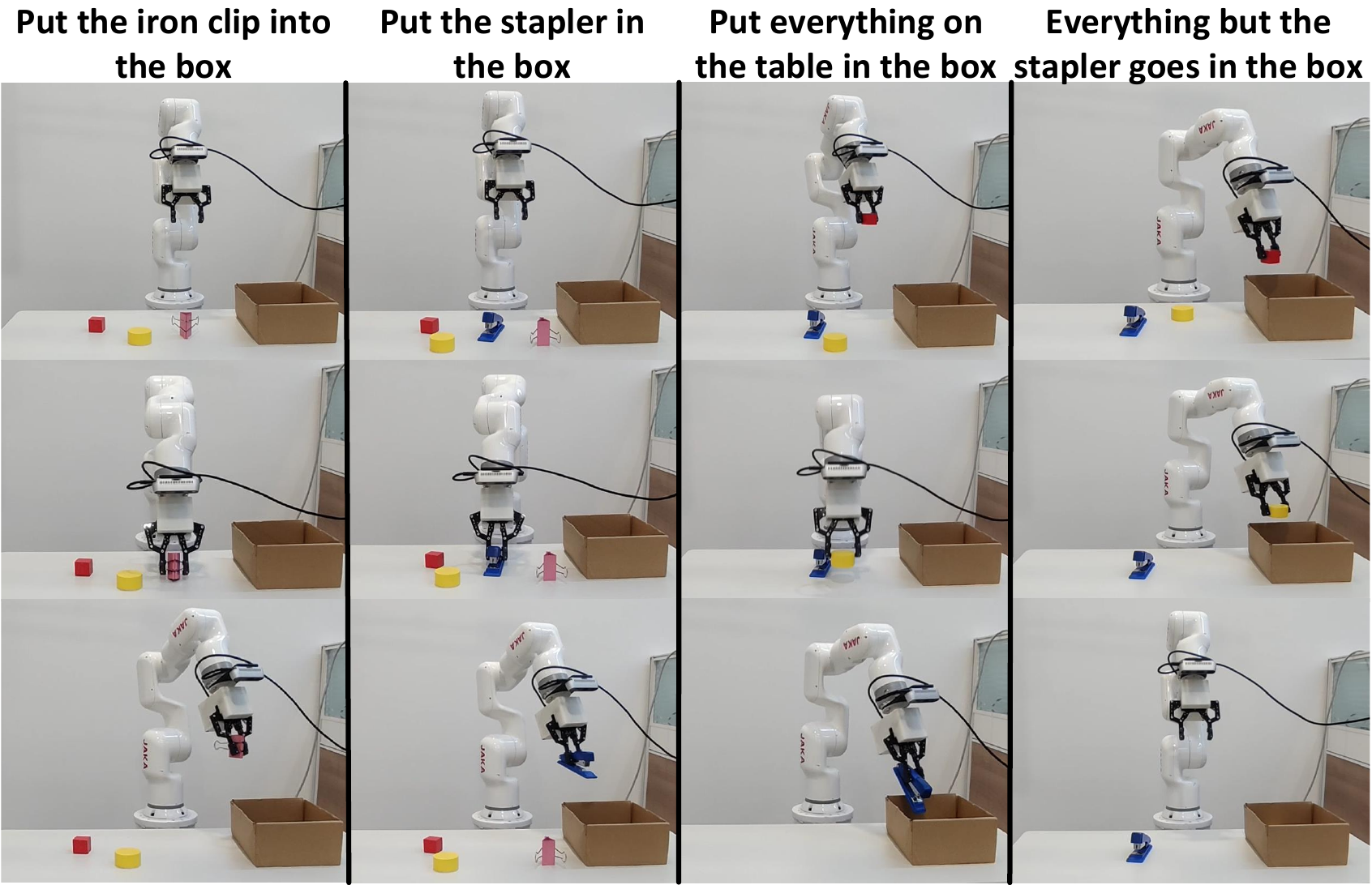}
\caption{
Task execution in four real-world scenarios, with each column representing one task. From left to right, the images illustrate the execution process of the four tasks listed in \cref{table:real_data}.
}
\label{fig:4real_task}
\end{figure}

\subsubsection{Experiment Setup}
In the real-world experiments, we employed a JAKA Mini 2 robotic arm and an Intel RealSense D435i RGB-D camera. Four tabletop manipulation tasks were designed, covering both high-level and low-level natural language instructions. An egocentric camera mounted on the robotic arm captured RGB observations at a resolution of 640×480.

We implemented three challenging tasks using low-level natural language commands involving everyday objects, along with one task using a high-level instruction. These tasks included:
\begin{enumerate*}[label=\arabic*.]
\item Placing a metal clip into a cardboard box.
\item Placing a stapler into a cardboard box.
\item Putting all objects on the table into the box.
\item Everything but the stapler goes in the box.
\end{enumerate*}
All tasks were unseen during training and were not included in any of the agent’s prompts.
In the task of placing a metal clip into a cardboard box, we deliberately avoided specifying the clip’s color (e.g., “pink metal clip”) to ensure that the agent relied on true multimodal perception of physical object properties, rather than relying on shortcut cues based on color features.
In the tabletop cleanup task, we added a black desk mat to the scene to test the agent’s performance in a more visually complex environment.

\begin{table}[htbp]
\centering
\caption{Experimental Results of PFEA in Four Real-World tasks}
\label{table:real_data}
\begin{tabular}{lcc} 
\toprule
                                           & \multicolumn{2}{c}{Success Rate}  \\ 
\cmidrule{2-3}
Tasks                                      & PFEA          & LLM               \\ 
\midrule
Put the iron clip into
the box             & \textbf{80\%} & 55\%              \\
Put the stapler in the
box                 & \textbf{65\%} & 50\%              \\
Put
everything on the table in the box     & \textbf{70\%} & 30\%              \\
Everything
but the stapler goes in the box & \textbf{60\%} & 25\%              \\ 
\midrule
Total~                                     & \textbf{68\%} & 40\%              \\
\bottomrule
\end{tabular}
\end{table}

\subsubsection{Real-world Task Execution}
The agent demonstrated strong task performance in real-world scenarios, as illustrated in Figure~\ref{fig:4real_task}.
In evaluations across four real-world task scenarios, our agent achieved a 28\% higher success rate compared to approaches that rely solely on LLMs for performing high-level natural language tasks. This significant improvement is attributed to the agent’s effective design in task planning and feedback evaluation mechanisms. Detailed experimental results are presented in Table~\ref{table:real_data}.
Each task was tested 20 times.
Experiments conducted in real-world scenarios further validate the effectiveness of our agent in practical applications. Through diverse testing environments and the challenges posed by real settings, the agent demonstrates strong adaptability, clearly highlighting the feasibility and advantages of our framework for human-centered tasks in real-world deployments.

\subsection{Ablation Study and Analysis}

\begin{table}[htbp]
\centering
\caption{Results of the Comparative Experiments across 20 Tasks}
\label{table:three_data}
\begin{tabular}{lccc} 
\toprule
                     & \multicolumn{3}{c}{Average Success Rate} \\ 
\cmidrule{2-4}
                     & \makecell{PFEA \\ w/o Planner} 
                     & \makecell{PFEA \\ w/o Evaluator} 
                     & PFEA \\ 
\midrule
Ten Prompted Tasks   & 69\%                 & 76\%                   &\textbf{82\%} \\
Ten Unprompted Tasks & 45\%                 & 66\%                   &\textbf{74\%} \\ 
\bottomrule
\end{tabular}
\end{table}

\subsubsection{Ablation Study}
In the ablation study, we designed two control settings: one with the planner removed and the other with the evaluator removed. The experimental tasks were identical to those in Table~\ref{table:sim_data}, comprising a total of 20 tasks. Unlike the comparative experiments, the success rate calculation was adjusted in the ablation study. Specifically, in the comparative experiments, each task was executed 20 times, with a success recorded as 1 and a failure as 0. In contrast, in the ablation study, execution of the corresponding action was assigned a score of 0.25, full completion of the task requirements was scored as 1, and failure was scored as 0. For each task, the 20 execution results were summed and averaged, after which an overall mean was obtained across 10 tasks. The results of the ablation study are presented in Table~\ref{table:three_data} (w/o = without).

\subsubsection{Analysis}
Since all hierarchical methods exhibit limitations in the execution strategies of low-level steps, task failures may arise either from planning errors or from shortcomings in low-level control policies. Experimental observations indicate that the discrepancy between planning and execution success rates primarily stems from the following factors: minor inaccuracies in position recognition, deviations in the two-finger gripper at the end of the robotic arm, and disturbances caused by friction between the gripper and the object. Notably, in real-world scenarios, friction exerts a substantial influence on the grasping process. In addition, the grasping angle introduces further variability. For instance, when grasping a metal clip, the success rate differs depending on whether the front or the rear end of the clip is targeted.

\section{Conclusions}
In this paper, we propose a novel LLM-based vision and language agent framework for robots operating in the physical world. This framework enables robots to understand and execute human-centered, high-level natural language instructions, thereby completing complex real-world tasks.
First, we introduce a speech recognition and synthesis module that enables bidirectional conversion between human speech commands and text, achieving natural human-robot communication.
Second, we design an agent task planner with visual environmental perception capabilities. Leveraging LLMs, this planner decomposes high-level human natural language instructions into multi-step executable commands for the robot.
Furthermore, we construct an agent converter that converts specific natural language instructions into directly executable Python control code.
We also design an agent evaluator equipped with assessment and feedback mechanisms, which are crucial for completing complex tasks.
Finally, we implement an action execution module on both simulated and real robotic platforms, using an open-vocabulary model to identify object locations, enabling object grasping and placement.
Experimental results demonstrate that the proposed framework effectively enables robots to perform human-centered tasks using high-level natural language instructions in both simulated and real-world environments.
This work contributes toward the future development of an agent model context protocol that spans online and offline, virtual and physical environments, ultimately aiming to realize human-centered industrial intelligence.
The main limitation of this work lies in the agent’s limited ability to interpret low-level ambiguous instructions. 
Future work will explore the matching mechanisms between ambiguous language and visual input using VLMs, and integrate them into our agent framework to further enhance its capacity to execute human-centered tasks.

\bibliographystyle{IEEEtran}

\end{document}